\documentclass[times,twocolumn,final]{elsarticle}

\usepackage{cag}
\usepackage{framed,multirow}

\usepackage{amssymb}
\usepackage{latexsym}

\usepackage{lipsum}  

\usepackage{times}
\usepackage{epsfig}
\usepackage{graphicx}
\usepackage{amsmath}
\usepackage{amssymb}

\usepackage{booktabs}
\usepackage{multirow}
\usepackage{amsmath}
\usepackage{float} 
\usepackage{amsfonts}
\usepackage{enumitem}
\usepackage{subcaption}
\usepackage{lipsum}  
\definecolor{newcolor}{rgb}{.8,.349,.1}

\usepackage{hyperref}
\usepackage{duckuments}

\usepackage{amsmath}

\DeclareMathOperator*{\argmin}{arg\,min}

\definecolor{amethyst}{rgb}{0.6, 0.4, 0.8}
\definecolor{darkpastelgreen}{rgb}{0.01, 0.75, 0.24}
\definecolor{amber}{rgb}{1.0, 0.75, 0.0}
\definecolor{cadmiumorange}{rgb}{0.93, 0.53, 0.18}
\definecolor{lawngreen}{rgb}{0.49, 0.99, 0.0}
\definecolor{limegreen}{rgb}{0.2, 0.8, 0.2}
\definecolor{neongreen}{rgb}{0.22, 0.88, 0.08}
\definecolor{amethyst}{rgb}{0.6, 0.4, 0.8}
\definecolor{darkpastelgreen}{rgb}{0.01, 0.75, 0.24}
\definecolor{greenbest}{RGB}{88,137,15}
\definecolor{redworst}{RGB}{137,15,27}

\newcommand{\REMOVE}[1]{{}}

\newcommand{\Carlos}[1]{\textcolor{olive}{[\textbf{Carlos}: {#1}]}}
\newcommand{\Elena}[1]{\textcolor{magenta}{[\textbf{Elena}: {#1}]}}

\newcommand{\etal}{\emph{et~al.}}

\newcommand{\camera}{\omega_o}
\newcommand{\light}{\omega_i}

\usepackage{url}
\usepackage{xcolor}
\definecolor{newcolor}{rgb}{.8,.349,.1}

\usepackage{hyperref}

\usepackage[switch]{lineno} %

\journal{Computers \& Graphics}

\begin{document}

\verso{Preprint Submitted for CEIG'23}
 
\begin{frontmatter}

\title{NeuBTF: Neural fields for BTF encoding and transfer}

\author[1]{Carlos \snm{Rodriguez-Pardo}\corref{cor1}}
\cortext[cor1]{Corresponding author}
\emailauthor{carlos.rodriguezpardo.jimenez@gmail.com}{Carlos Rodriguez-Pardo}

\author[2]{Konstantinos \snm{Kazatzis}}

\author[3]{Jorge \snm{Lopez-Moreno}}

\author[4]{Elena \snm{Garces}}

\address[1,3,4]{SEDDI and Universidad Rey Juan Carlos}
\address[2]{SEDDI}

\received{\today}

\begin{abstract}
Neural material representations are becoming a popular way to represent materials for rendering. They are more expressive than analytic models and occupy less memory than tabulated BTFs. However, existing neural materials are immutable, meaning that their output for a certain query of UVs, camera, and light vector is fixed once they are trained. 
While this is practical when there is no need to edit the material, it can become very limiting when the fragment of the material used for training is too small or not tileable, which frequently happens when the material has been captured with a gonioreflectometer. 
In this paper, we propose a novel neural material representation which jointly tackles the problems of BTF compression, tiling, and extrapolation. 
At test time, our method uses a guidance image as input to condition the neural BTF to the structural features of this input image. Then, the neural BTF can be queried as a regular BTF using UVs, camera, and light vectors. 
Every component in our framework is purposefully designed to maximize BTF encoding quality at minimal parameter count and computational complexity,
achieving competitive compression rates compared with previous work.
We demonstrate the results of our method on a variety of synthetic and captured materials, showing its generality and capacity to learn to represent many optical properties.
\end{abstract}

\begin{keyword}
\KWD Neural Fields \sep Reflectance \sep Rendering \sep BTF Compression
\end{keyword}

\end{frontmatter}

\section{Introduction}\label{sec:introduction}

A common approach to modeling real-world spatially-varying materials in computer graphics is through the use of Bidirectional Texture Functions (BTFs). This type of representation models the dense optical response of the material, and is more general than analytic representations such as microfacet SVBRDF. However, BTFs can occupy large amounts of memory.
Recently, neural material representations are being proposed as a learning-based alternative to tabulated BTFs, providing a more compact solution while keeping the flexibility and generality of BTFs.

Creating digital representations of real material samples requires using an optical capture device, such as a gonioreflectometer, %
a smartphone~\cite{guo2020materialgan}, or a flatbed scanner~\cite{rodriguezpardo2023UMat}. During the process, several choices must be made. First, it is important to select a patch of the material that contains enough spatial variability. Second, a process --automatic or manual-- must be found to produce a tileable material that can be used to create seamless 3D renders. Finally, resources must be allocated for storage as needed.
Making these choices when dealing with implicit or tabulated representations, such as in BTFs or neural materials, is particularly crucial. Once these representations are trained or captured, they cannot be easily modified and it is only possible to query them using the UVs, light, and camera vectors.

In this paper, we propose a novel neural material representation that addresses these issues. Unlike existing neural approaches that are immutable once trained~\cite{rainer2019neural,kuznetsov2021neumip,rainer2020unified,kuznetsov2022rendering}, our model can be queried at test time with a guidance image that conditions the neural BTF to the structure provided by the guidance image. Our approach resembles synthesis by example and procedural processes, and can be used to extrapolate BTFs to large material samples, as well as to easily create tileable ones. Furthermore, our method achieves better compression rates than previous work on neural BTF representations.

To achieve this, we present a novel method that works at two steps.  
In the first step, we condition the neural BTF using a \emph{guidance image} as input. To this end, we use an \emph{autoencoder} that outputs a high-dimensional latent representation of the material, a \emph{neural texture}, which jointly encodes reflectance and structural properties. In the second step, the UV position of the latent representation, along with the camera and light vectors,  are decoded by a fully-convolutional sinusoidal decoder, a neural \emph{renderer} to obtain the RGB values. %
Using a single BTF as input, we train the network end-to-end using a custom training procedure, loss function, and data augmentation policy. This policy, inspired by recent work on attribute transfer~\cite{rodriguezpardo2021transfer}, allows the autoencoder to encode the relationship between structural features and reflectance, enabling the propagation of the BTF to novel input guidances.
Once trained, the novel input guidances may come from the same material, a different material, or a structural pattern. An input guidance of the same material can be used to extrapolate the BTF to larger samples or create tileable BTFs, provided the input guidance is tileable. If the input guidace is a structural pattern, the local features can be used to synthesize novel materials.

In summary, we propose the following contributions:
\begin{itemize}
	\item The first neural BTF representation with conditional input that can be used to extrapolate BTF measurements, easily create tileable BTFs, and synthesize novel materials.
	\item We show how to leverage our system for rendering large-scale and tileable neural BTF generation using measurements captured with small portions of the material. 
	\item We demonstrate that our method works with synthetic and captured materials of diverse optical properties, including colored specular or anisotropy. 
\end{itemize}

We provide additional results, supplementary materials, and implementation details at \href{https://carlosrodriguezpardo.es/projects/NeuBTF/}{our project website}.

\section{Related Work}\label{sec:relatedwork}

An accurate method for representing the optical properties of materials is through Bidirectional Texture Functions (BTFs) \cite{dana2001brdf}. BTFs are 6D functions that characterize all possible combinations of incoming and outgoing light and camera directions for the 2D spatial extent of a material. 
Although they are successful in representing materials, they have a major drawback in terms of memory requirements. Therefore, BTF compression has been a major research topic~\cite{filip2008bidirectional}.
Non-neural approaches used dimensionality reduction techniques such as Principal Component Analysis (PCA)~\cite{koudelka2003acquisition,muller2003compression,weinmann2014material}, vector quantization~\cite{havran2010bidirectional}, or clustering~\cite{tong2002synthesis}. However, these approaches were recently surpassed by neural models~\cite{hinton2006reducing} due to their flexibility and superior capacity to learn non-linear functions.  

\paragraph*{Neural BTFs}
Rainer \emph{et~al.} \cite{rainer2019neural} proposed the first method to use deep autoencoders to compress BTFs, surpassing PCA~\cite{weinmann2014material} on captured BTFs. However, this approach required training a single neural network per material. To address this limitation, a later work by the same authors~\cite{rainer2020unified} proposed a generalization of this idea in which a single network was able to generalize to a variety of materials.
Although these methods were very effective for compressing flat materials, they had some limitations when it came to modeling materials with volume. In their work, Kuznetsov~\emph{et~al.} \cite{kuznetsov2021neumip} improved the quality of neural materials by introducing a neural offset module that captures parallax effects. Further, they method also allowed for level-of-detail though MIP mapping by training a multi-resolution neural representation.
However, grazing angles and silhouette effects remained a challenge for this approach. In a subsequent work, Kuznetsov~\emph{et~al.}~\cite{kuznetsov2022rendering} explicitly trained the network using queries that span surface curvatures, effectively handling these cases.
Representing fur, fabrics, and grass with neural reflectance fields was explored by Baatz~\emph{et~al.}~\cite{baatz2022nerf} who proposed a representation that jointly models reflectance and geometry.

All of these approaches share the idea of querying neural material using UVs, camera, and lighting vectors, but  do not provide any functionality for modifying the material once the network is trained. In contrast, our approach can take a guidance image as input, which conditions the output to generate material variability.

\begin{figure*}[ht!]
	\centering
	\includegraphics[width=1\linewidth]{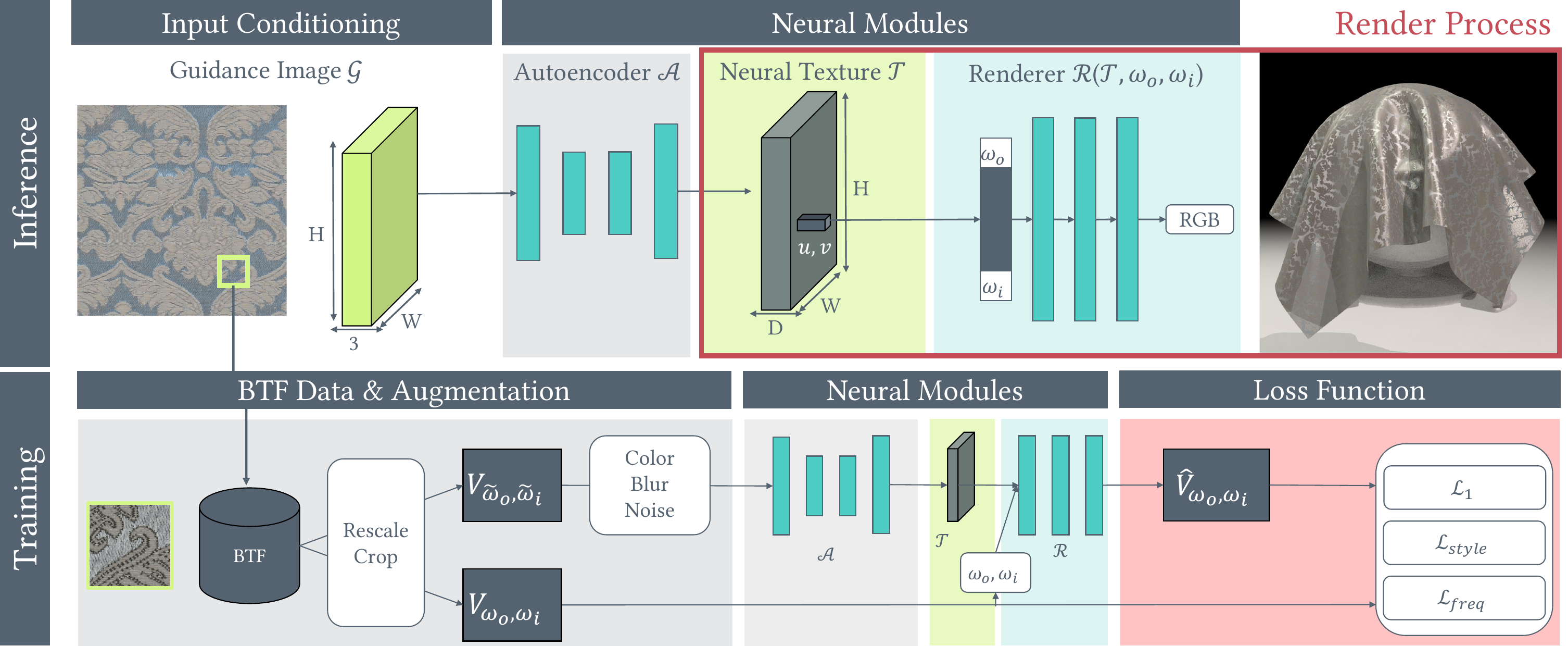}
	\caption{An overview of our neural BTF inference and training processes. Top-Inference: Using a guidance image $\mathcal{G} \in \mathbb{R}^{H \times W \times 3}$, we use our trained autoencoder to generate the Neural Texture $\mathcal{A}(\mathcal{G}) = \mathcal{T}_\mathcal{G} \in \mathbb{R}^{H \times W \times D}$, which preserves the spatial resolution of the input image but represents a higher-dimensional learned representation. This Neural Texture $\mathcal{T}_\mathcal{G}$, along with the trained renderer $\mathcal{R}$, can be queried as a regular BTF, using UVs, and target camera and light positions for regular rendering. 
		Bottom-Train: During training, following previous work on photometric data augmentation~\cite{rodriguezpardo2021transfer}, we randomly select an input \emph{view} 
		$V_{\tilde{\omega}_o, \tilde{\omega}_i}$ and a target $V_{\omega_o, \omega_i}$. This allows the model to generalize to novel light or camera conditions and acts as a regularizer.  To both views, we apply random rescale and cropping. Then, only to $\text{V}_{\tilde{\omega}_o, \tilde{\omega}_i}$, we randomly apply hue variations, gaussian blur, and noise, and feed it to the \emph{autoencoder}, which returns a 2D \emph{latent representation} of the material. A fully-convolutional \emph{decoder} with sinusoidal activations receives both this latent space and the target $\camera, \light$ camera and light angles, and estimates $\hat{V}_{\camera, \light}$. This output is compared with $V_{\camera, \light}$ using a multifaceted loss function.}
	\label{fig:overview2}
\end{figure*}

\paragraph*{Material synthesis and tiling}  

Texture synthesis is a long-standing problem in the field of computer graphics. The goal is to reconstruct a larger image given a small sample, leveraging the structural content and internal statistics of the input image. This concept has been used for synthesizing single images, BTFs, and full material models. For images, the most common strategies include PatchMatch \cite{diamanti2015synthesis}, texture transport~\cite{aittala2015two}, point processes~\cite{guehl2020semi,lefebvre2006appearance}, or neural networks~\cite{elad2017style, zhou2018non, fruhstuck2019tilegan,rodriguezpardo2022SeamlessGAN}.
BTF synthesis, however, has received less attention. Steinhausen et al.~\cite{steinhausen2015crossdevice,steinhausen2015normals} extrapolated BTF captures to larger material samples using non-neural texture synthesis methods. For full materials, Li~\etal~\cite{lin2019site} captured the appearance of materials by first estimating their BRDF and then synthesizing the high-resolution micro-structure from a dataset of measured SVBRDFs. Nagano~\etal~\cite{nagano2015skin} measured microscopic patches of the skin and used a convolutional filter to propagate the measurements to a spatially-varying texture. Deschaintre~\etal~\cite{deschaintre2020guided} used an autoencoder to propagate SVBRDFs to large material samples. Also recently, Rodriguez-Pardo and Garces~\cite{rodriguezpardo2021transfer} propagated any kind of visual attribute having a single image as guidance. Their approach shares some similarity to ours, although they transfer 2D image attributes, while we transfer the full BTF.

Procedural models~\cite{hu2022controlling,zhou2023semi,zhou2022tilegen} are nowadays very successful for generating tileable materials. Thanks to the use of a tileable template, these methods adjust the generated image to the features available in the template. As we show, our approach can also work with a binary template as input. However, guaranteeing predictable outputs given this kind of input is out of the scope of our technique, which can transfer BTF measurements having as input a guidance image of the same material. %

\paragraph*{Other neural representations in rendering}

Limited to BRDFs, neural networks trained with adaptive angular sampling have been explored to enable importance sampling~\cite{sztrajman2021neural}, needed for Monte Carlo integration. Deep latent representations also allow for BRDF editions. For example, Hu~\emph{et~al.}~\cite{hu2020deepbrdf} demonstrate that autoencoders can outperform classic PCA for the purpose of editing. 
  Other applications of neural encodings in rendering are numerous. For instance, they have been used for scene prefiltering~\cite{bako2023deep}, where geometry and materials are simplified to accommodate the LoD of the scene using a voxel-based representation and trained latent encodings. For anisotropic microfacets, Gauthier~\emph{et~al.}~\cite{gauthier2022mipnet} propose a cascaded architecture able to adjust the material parameters to the MIP mapping level. Encoding light transport using neural networks for real-time global illumination has also been explored~\cite{rainer2022neural,gao2022neural}, showcasing promising results.

\section{Method}\label{sec:overview}

\begin{figure*}[ht!]
	\centering
	\includegraphics[width=1\linewidth]{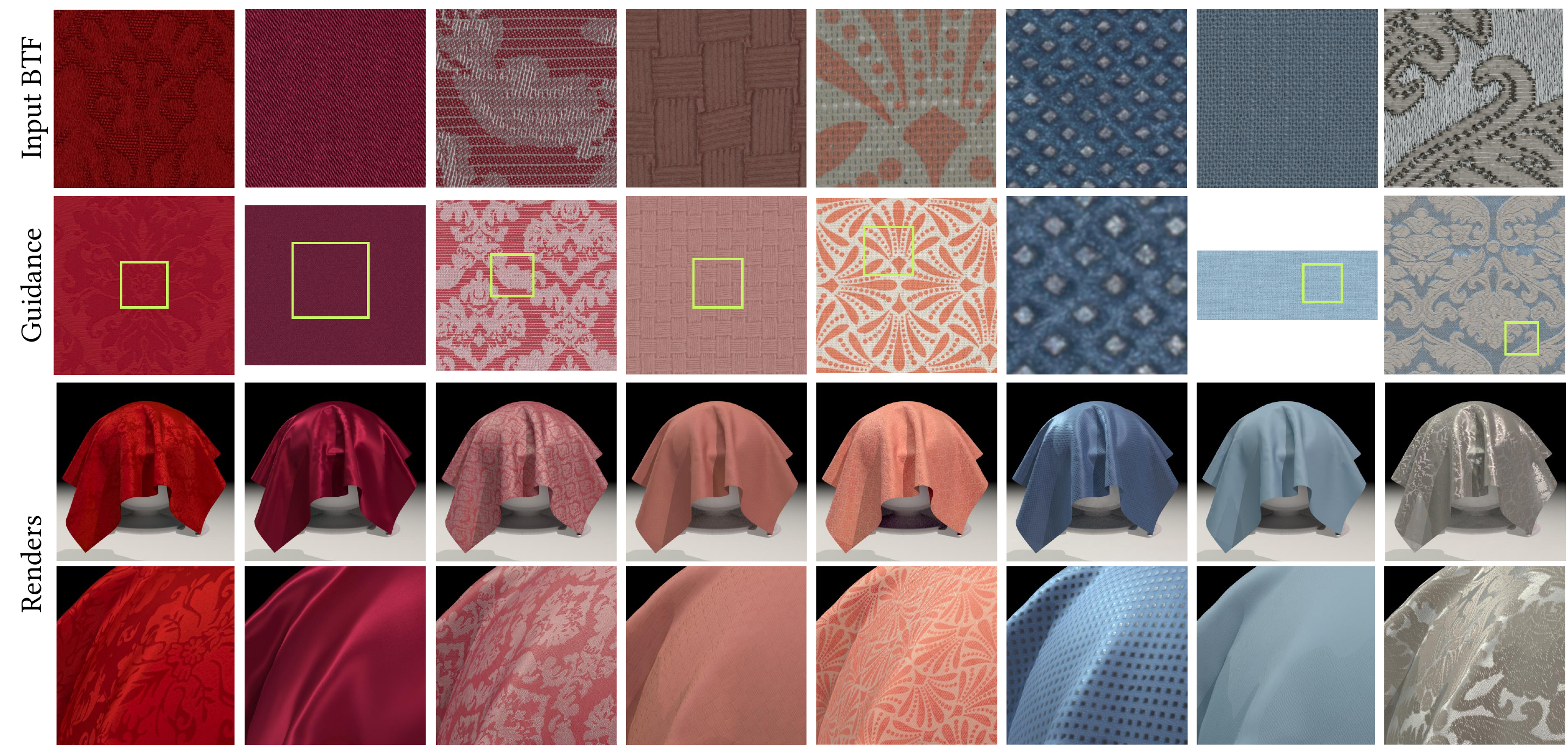}
	\caption{Some tileable neural materials achieved with our method. On the top row, we show a slice of the BTF used to train a NeuBTF representation. With the tileable guidance images shown on the second row, we propagate the neural texture using our autoencoders. These neural textures can be rendered to generate realistic images (third). We provide closeups on the bottom row. In the cases where the guidance image covers a larger area than the training crop, we highlight the training surface area as a green inset.}
	\label{fig:tiling2}
\end{figure*}

We present an overview of our approach in Figure~\ref{fig:overview2}, where we show our inference and training pipelines. Our goal is twofold: First, find a compact representation for a BTF through the use of neural networks. Second, enable the extrapolation of the BTF according to guidance images used as input. In Section~\ref{sec:inference} we describe our inference pipeline and neural network,  and in Section~\ref{sec:training} our training process. Section~\ref{sec:model_design} contains specific implementation details and design of the neural networks. 

\subsection{Inference} \label{sec:inference}

Our neural network is composed of three modules: an \textit{autoencoder} $\mathcal{A}$, a \emph{neural texture} $\mathcal{T} \in \mathbb{R}^{H \times W \times D}$, and a \emph{renderer} $\mathcal{R}$. 
The renderer $\mathcal{R}(\mathcal{T}(\text{u},\text{v}), \camera, \light) = \text{RGB}$ takes as input the feature vector at the $(\text{u}, \text{v})$ coordinates of the neural texture $\mathcal{T}$, the view $\camera$ and light $\light$ positions, and returns an RGB value. $\mathcal{R}$ acts as a conventional BTF and can be used as such in any render engine.

An input \textit{guidance} image, $\mathcal{G}\in \mathbb{R}^{H \times W \times 3}$, is used during inference to condition the generation of the \emph{neural texture} $\mathcal{T}$. 
This conditioning allows us to propagate the learned reflectance to novel guidance images that can be: a larger sample of the same material, a different material, or a structural image. In the simpler case, the guidance image comes from the BTF used for training, and our process is equivalent to previous work~\cite{rainer2019neural,kuznetsov2021neumip}.

The autoencoder $\mathcal{A}$ takes the guidance image $\mathcal{G}\in \mathbb{R}^{H \times W \times 3}$ as input and outputs a neural texture $\mathcal{T} \in \mathbb{R}^{H \times W \times D}$ with the same size $H \times W$ as the input guidance image but with more latent dimensions $D$. As a result, each pixel in the guidance image has a higher-dimensional neural representation in $\mathcal{T}$. Because it is trained without explicit supervision, this latent representation can capture the reflectance and structural patterns automatically. 
Kuznetsov~\etal~\cite{kuznetsov2021neumip} also used a latent neural representation of the material, however, lacking the initial autoencoder their approach cannot synthesize novel BTFs without retraining, while our conditioning module allow us to generate novel BTFs during test time. 
The autoencoder and the renderer are neural networks trained jointly, using an end-to-end image-to-image approach describe below.

\subsection{Training} \label{sec:training}

Figure~\ref{fig:overview2}~(bottom) illustrates our training process. It has two objectives: First, equivalent to regular BTF encoding, we aim to find the mapping between camera direction $\camera$, light direction $\light$, and output slices of the BTF: $(\camera, \light) \rightarrow \text{V}_{\camera, \light} \in \text{BTF}$. Second, we aim to condition the synthesis process with an input guidance image, $\mathcal{G}$. %
To this end, 
for training, we feed the model with images, $\text{V}_{\tilde{\omega}_o, \tilde{\omega}_i}$, which are randomly sampled from the BTF, and are subject to additional data augmentation processes. 
This extensive augmentation process guarantees invariance to different input variations during test time, like camera or illumination conditions, while keeps consistency of the outputs. 

\paragraph*{Loss Function Design}
Our loss function, that compares ground truth slices  $\text{V}_{\camera, \light}$ with generated ones  $\mathcal{R}(\mathcal{A}(f(\text{V}_{\tilde{\omega}_o, \tilde{\omega}_i})), \camera, \light)$, is a weighted sum of three terms: a pixel-wise loss, a  style loss, and a frequency loss,
\begin{equation}
	\mathcal{L} = \lambda_{\mathcal{L}_{1}} \mathcal{L}_{1} + \lambda_{style}   \mathcal{L}_{style} + \lambda_{freq}   \mathcal{L}_{freq}
\end{equation}
The main driver of our loss is the pixel-wise norm $\mathcal{L}_{1}$. $\mathcal{L}_1$ produces sharper results than higher-order alternatives, such as $\mathcal{L}_2$~\cite{isola2017image,rodriguezpardo2021transfer}.  Following~\cite{kuznetsov2021neumip}, we apply a $log(x+1)$ compression to improve the model results on high dynamic range. This compression is only done to the pixel-wise component of the loss function. Inspired by recent work on texture synthesis, capture and transfer~\cite{mardani2020neural,aittala2013practical,rodriguezpardo2022SeamlessGAN,rodriguezpardo2023UMat,zhou2022tilegen}, we introduce a $\mathcal{L}_{style}$ loss to help the model generate higher quality and sharper results. Further, to mitigate the spectral bias of convolutional neural networks and help ameliorate the results further, we also introduce a \emph{focal frequency loss} into our learning framework~\cite{jiang2021focal}. This combination of loss functions proves effective for our problem, without the need for complex adversarial losses which could reduce efficiency or destabilize training.

\REMOVE{In this section, we describe our novel neural representation for material reflectance (Sec.~\ref{sec:material_model}),  discuss our training procedure  (Sec.~\ref{sec:model_design}), model design (Sec.~\ref{sec:training}), and the inference process (Sec.~\ref{sec:inference}). We illustrate our material model training in Figure~\ref{fig:overview}, and the inference process in Figure~\ref{fig:training_inference}. Implementation details are provided on Section~\ref{sec:implementation} and extended on the supplementary material.}

\REMOVE{ 
Our training objective is to minimize a loss function $\mathcal{L}$ by learning to map from input $\textsc{BTF}_{c_i, l_i}$ to target $V_{c_t, l_t}$ images, as follows:
\begin{equation}
	\argmin_{\mathcal{A},\mathcal{R}} \mathcal{L}\left(V_{c_t, l_t}, \mathcal{R}(\mathcal{A}(V_{c_i, l_i}), c_t, l_t)\right) \forall c \in C, l \in L
\end{equation}

Where $C$ and $L$ are the set of available camera and lighting positions of the input BTF, respectively. Note that this optimization objective does not explicitly optimize the Neural Texture $\mathcal{T}$. Its values are learned without supervision, they are found automatically. $\mathcal{T}$ must have values that are both easy to transfer and which can be effectively used to represent the reflectance of the material, decoded by the renderer $\mathcal{R}$. This double optimization objective shapes the type of properties which can be represented in $\mathcal{T}$, which we explore qualitatively on Section~\ref{sec:evaluation}. \Carlos{Show some examples of $\mathcal{T}$ for a couple of materials.} We design the loss function, training procedure and the architecture of $\mathcal{A}$ and $\mathcal{R}$ to maximize the reconstruction quality of the input BTF, to enable high-quality propagations at test time, and to maintain a low parameter count, maximizing compression. }

\paragraph*{Data Augmentation}
We train our models using a comprehensive data augmentation policy aimed at achieving high quality reflectance propagation, increasing performance and generalization, and allowing for generation of multiple resolution materials at test time. We build upon recent work on material transfer~\cite{rodriguezpardo2021transfer} and use images of the material taken under different illumination and viewing conditions as inputs to our autoencoder. This helps it generalize to novel capture setups, which allows for multiple applications we describe on Section~\ref{sec:applications}. In particular, we use every image available on the input BTF, selected uniformly at random for each element in each batch during training. 
As in~\cite{rodriguezpardo2021transfer}, we also use random rescaling, which helps the model generalize to new scales, and build neural materials of multiple resolutions at test time, as we describe on Section~\ref{sec:lod}. 
Inspired by recent work on image synthesis~\cite{rodriguezpardo2021transfer,Texler20-SIG,rodriguezpardo2022SeamlessGAN}, we use random cropping, which helps generalization by effectively increasing the dataset size. Finally, we extend the color augmentation policy in~\cite{rodriguezpardo2021transfer} with random hue changes across the entire color wheel, and introduce random Gaussian noise and blurs to the input images, to help it generalize further, as proposed in~\cite{rodriguezpardo2023UMat}.

\REMOVE{
\paragraph*{Loss Function Design}
Our loss function is a weighted sum of three terms: a pixel-wise loss, a  style loss, and a frequency loss:

\begin{equation}
	\mathcal{L} = \lambda_{\mathcal{L}_{1}} \mathcal{L}_{1} + \lambda_{style}   \mathcal{L}_{style} + \lambda_{freq}   \mathcal{L}_{freq}
\end{equation}

The main driver of our loss is the pixel-wise norm $\mathcal{L}_{1}$. $\mathcal{L}_1$ produces sharper results than higher-order alternatives, such as $\mathcal{L}_2$~\cite{isola2017image,rodriguezpardo2021transfer}.  Following~\cite{kuznetsov2021neumip}, we apply a $log(x+1)$ compression to improve the model results on high dynamic range. This compression is only done to the pixel-wise component of the loss function. Inspired by recent work on texture synthesis, capture and transfer~\cite{mardani2020neural,aittala2013practical,rodriguezpardo2022SeamlessGAN,rodriguezpardo2023UMat,zhou2022tilegen}, we introduce a $\mathcal{L}_{style}$ loss to help the model generate higher quality and sharper results. Further, to mitigate the spectral bias of convolutional neural networks and help ameliorate the results further, we also introduce a \emph{focal frequency loss} into our learning framework~\cite{jiang2021focal}. This combination of loss functions proves effective for our problem, without the need for complex adversarial losses which could reduce efficiency or destabilize training. 
}

\REMOVE{

\subsection{Inference} \label{sec:inference}
Once the model has been trained, we can generate the final Neural BTF. To do so, we require a guidance image $\mathcal{G}\in \mathbb{R}^{3\times H \times W}$ as input to the autoencoder $\mathcal{A}$. $\mathcal{G}$ can take many forms: it can an image of the input BTF, an larger portion of the same material where more variations are represented, a tileable texture, or even a synthetic image for which we want to propagate reflectance values, as shown in Figure~\ref{fig:propagation}. From it, we generate the Neural Texture $\mathcal{A}(\mathcal{G}) = \mathcal{T}_\mathcal{G} \in \mathbb{R}^{D\times H \times W}$. 

$\mathcal{T}_\mathcal{G}$ and the decoder $\mathcal{R}$ form the final Neural BTF, which can then be queried as a regular BTF, using input UVs, camera and light vectors. We illustrate the generation of $\mathcal{T}_\mathcal{G}$ and the usage of the Neural BTF on rendering scenarios on Figure~\ref{fig:training_inference}.
}

\begin{figure}[t!]
	\centering
	\includegraphics[width=1\linewidth]{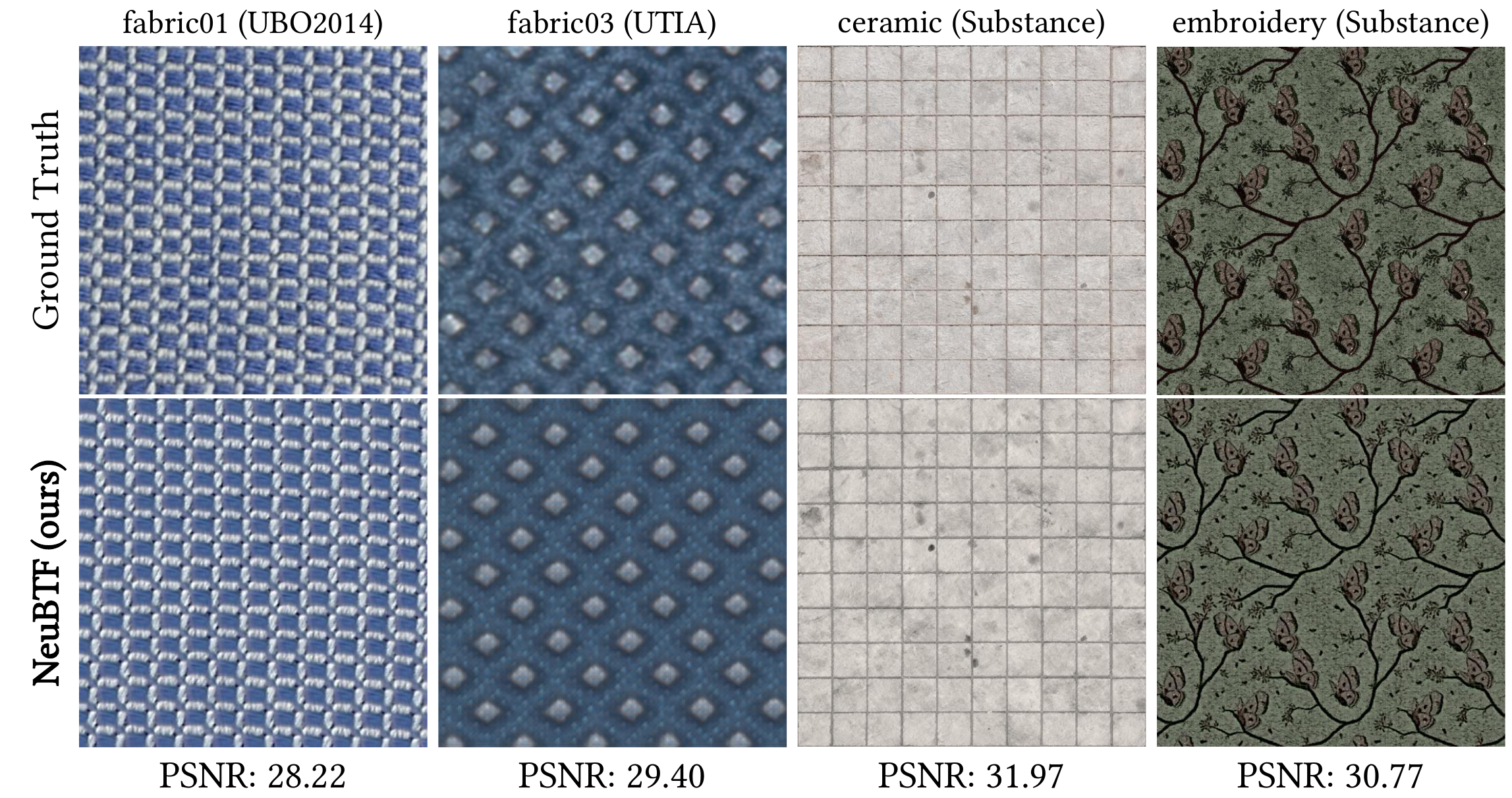}
	\caption{Qualitative results on a variety of BTFs, from different sources. From left to right, we show results on a material from the UBO~\cite{weinmann2014material} BTF dataset, the UTIA~\cite{haindl12ERCIM} BTF dataset and two synthetic materials rendered from Substance SVBRDFs. }
	\label{fig:comparisons_qualitative}
\end{figure}

\section{Model Design and Implementation}
\label{sec:model_design}

We provide extensive implementation details for model sizes, training and data generation on the supplementary material.

\paragraph*{Autoencoder}For the \emph{autoencoder}, we use a lightweight U-Net~\cite{ronneberger2015u} with a few modifications to tailor it for our problem. Inspired by recent work on CNN design, we leverage ConvNext~\cite{liu2022convnet} Blocks across our model, with depth-wise convolutions using $5\times5$ kernels. We empirically observe that ConvNext blocks achieve higher quality structural editions at a lower parameter count than vanilla U-Net blocks. To further help convergence and preserve details in the input images, we use residual connections~\cite{he2016deep, diakogiannis2020resunet,rodriguezpardo2023UMat} in every convolutional block of the model. We use $1\times1$ convolutions on the skip connections and residual scaling~\cite{karras2020analyzing}. As in~\cite{liu2022convnet}, we use Layer Normalization~\cite{ba2016layer} and GELU non-linearities~\cite{hendrycks2016gaussian}.  On the bottleneck of the model, we introduce an attention module~\cite{woo2018cbam} to help the model learn longer-range dependencies. To avoid checkerboard artifacts~\cite{odena2016deconvolution}, we use nearest neighbors interpolation for upsampling. Inspired by recent work on tileable material generation~\cite{zhou2022tilegen}, we use \emph{circular padding} throughout the model. We initialize its weights using \emph{orthogonal initialization}~\cite{arpit2019initialize}, which helps avoiding exploding gradients. %

\paragraph*{Renderer} For the \emph{renderer}, we build upon SIREN~\cite{sitzmann2020implicit} MLPs, with additional modifications to enhance its performance for our problem. We use $1\times1$ convolutions instead of vanilla linear layers, to allow for end-to-end training using 2D images. Further, we introduce Layer Normalization~\cite{ba2016layer} before each sinusoidal non-linearity, which stabilizes training. Finally, inspired by~\cite{mehta2021modulated}, we use residual connections~\cite{he2016deep}, to help preserve the information of the input vector across the decoder layers. Model weight initialization follows~\cite{sitzmann2020implicit}. With sinusoidal activations, we observe significantly higher reconstruction quality and training dynamics than with ReLU~\cite{nair2010rectified} MLPs, which are common for BTF compression~\cite{rainer2019neural,rainer2020unified,kuznetsov2021neumip}. %
Because the network is fully-convolutional, it can take as input feature vectors of any size. This is very convenient for our use cases when the input guidance image have a size different from the size of the original BTF used to train it. The renderer can be evaluated very efficiently in GPU, at an average of $2.514 e^{-4} \pm 4.48 e^{-5}$ ms per sample.

\section{Evaluation}\label{sec:evaluation}

\REMOVE{
\subsection{Ablation Study}\label{sec:evaluation_ablation}

Aquí:
\begin{itemize}
    \item Estudio cualitativo de la influencia de SIREN, Loss Functions, RAdam, Orthogonal Normalization, Autoencoder Architecture, etc.
    \item Loss WRT number of layers in renderer, autoncoder, \# latent parameters, width of renderer, etc.
    \item quantitative study of these factors
    \item speed vs accuracy
    \item Use render aware losses like in UMat
\end{itemize}
}

\REMOVE{
\begin{figure*}[ht!]
	\centering
	\includegraphics[width=1\linewidth]{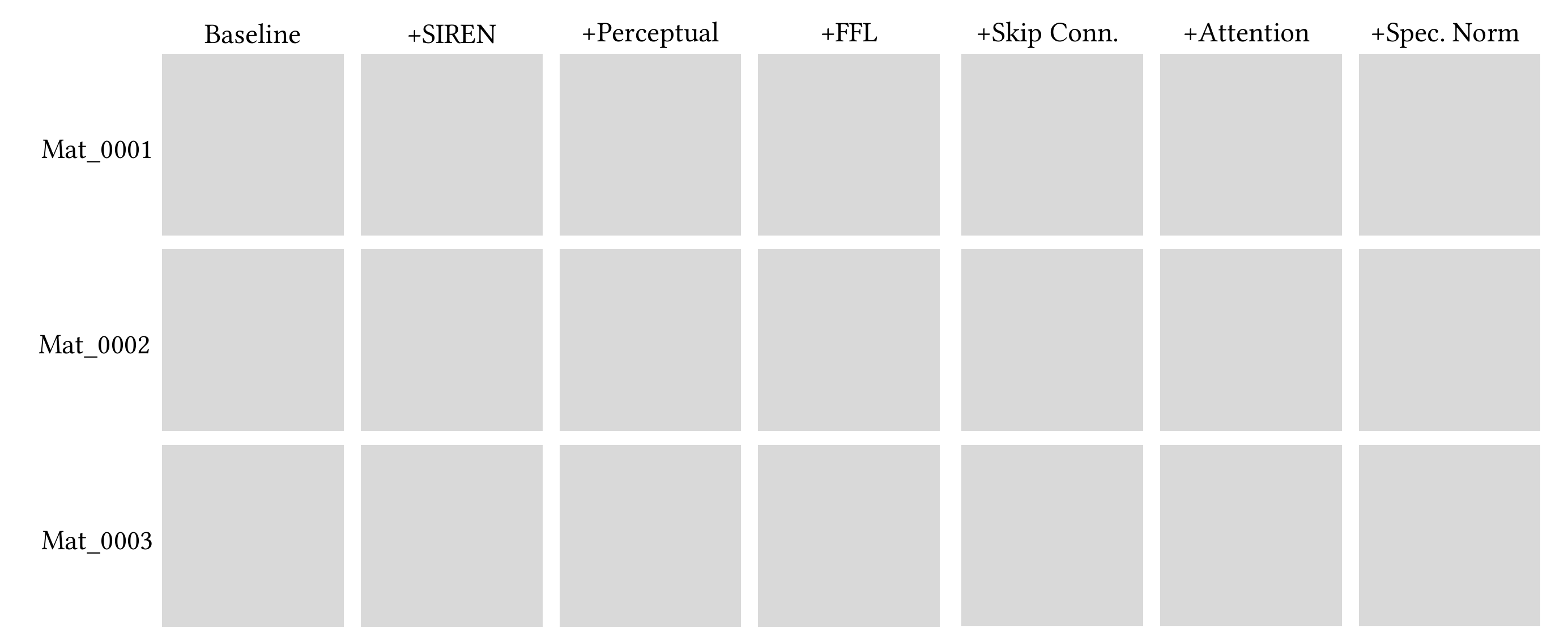}
	\caption{Qualitative results of our ablation study. From a baseline with a simple ReLU MLP and a baseline autoencoder, we progressively introduce additional modules into, in order, the MLP design, the loss function and the design of the encoder. Using a SIREN (cite) network, we achieve much higher reconstruction quality that with a simple MLP. Introducing perceptual and frequency losses help resolve additional details. Skip connections and attention help preserve the structure of the input images, while spectral normalization helps reduce the number of artifacts in the images. \Carlos{Candidato a irse}}
	\label{fig:ablation}
\end{figure*}}

\REMOVE{
\begin{itemize}
    \item Comparativa cuantitativa con Rainer2019 y Rainer2020 en materiales de Bonn y UTIA (intentar tenerlos)
    \item Comparativa cualitativa usando imágenes
    \item BRDF slices using BRDF viewer
\end{itemize}}

\begin{figure*}[t!]
	\centering
	\includegraphics[width=1\linewidth]{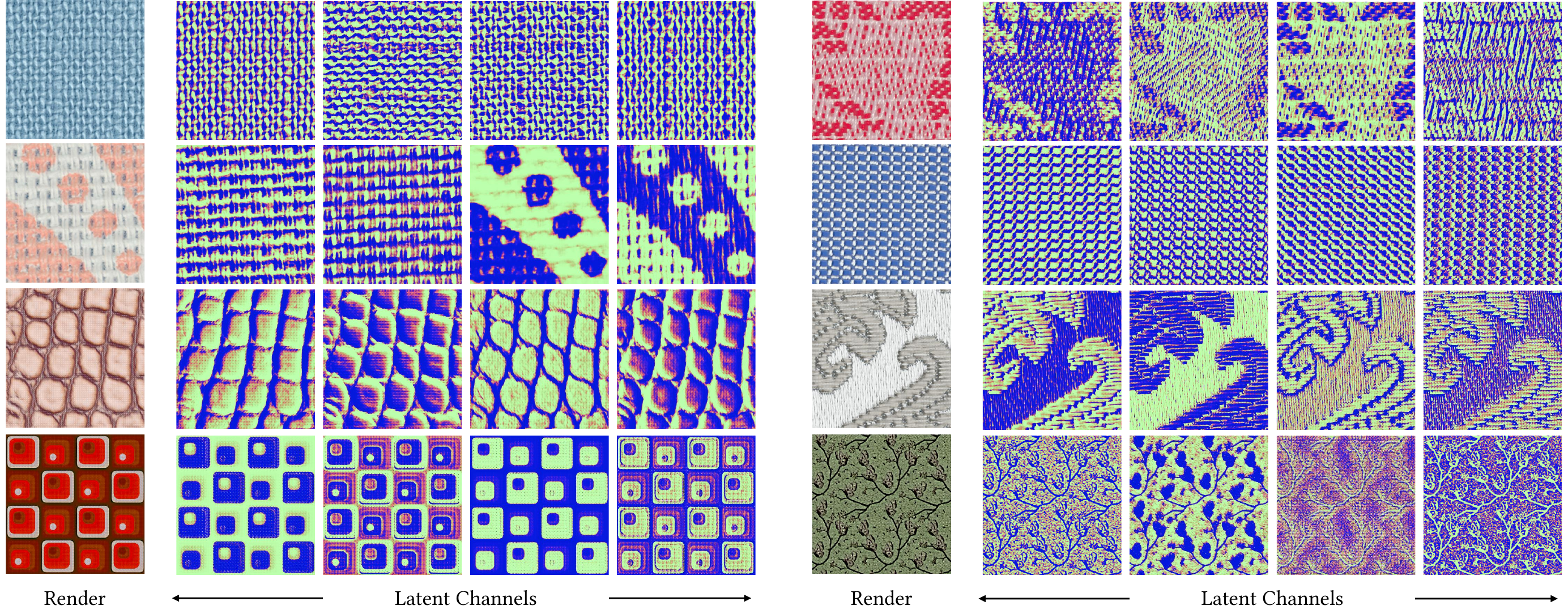}
	\vspace{-6mm}
	\caption{A selection of latent channels learned by NeuBTF for a variety of materials. We use a colorspace to help visualization. Without explicit supervision, the model internally learns semantically meaningful latent spaces. For instance, in the second example on the left, the two leftmost latent spaces encode geometry, while the other two encode the two distinct colors of the printed pattern over the yarns. }
	\label{fig:latent_channels}
\end{figure*}
\subsection{Qualitative and Quantitative Analysis}

\begin{table}[t!]
	\centering
	\resizebox{\columnwidth}{!}{%
		\begin{tabular}{@{}r|cccc@{}}
			\toprule
			\multicolumn{1}{r}{\textbf{Material}} & \textbf{fabric01}~\cite{weinmann2014material}  & \textbf{fabric03}~\cite{haindl12ERCIM}  & \textbf{ceramic}~(\emph{Subst.})                & \textbf{embroidery}~(\emph{Subst.})   \\ \midrule
			\multicolumn{1}{r}{PSNR $\uparrow$} & {26.58}$\pm$1.820 & 27.73$\pm$4.711     &29.19$\pm$2.111      & {28.79}$\pm$1.831  \\ 
			\multicolumn{1}{r}{SSIM~\cite{wang2004image} $\uparrow$} & {0.710}$\pm$0.108 &0.729$\pm$0.142& 0.819$\pm$0.110& {0.652}$\pm$0.156 \\ 
			\multicolumn{1}{r}{LPIPS~\cite{zhang2018unreasonable} $\downarrow$} &{0.451}$\pm$0.041   & 0.404$\pm$0.054 & 0.270$\pm$0.075& {0.341}$\pm$0.135  \\ 
			\multicolumn{1}{r}{FLIP~\cite{andersson2020flip} $\downarrow$} & {0.391}$\pm$0.047  &0.426$\pm$0.105  & 0.365$\pm$0.065  & {0.391}$\pm$0.115   \\ \bottomrule
		\end{tabular}%
	}
	\caption{Average ($\pm$ std.) reconstruction error across the full dimensional space for materials of different datasets, measured using pixel-wise and perceptual metrics.}
	\label{tab:Material}
\end{table}

We evaluate our method on materials from different sources including acquired BTFs from~\cite{weinmann2014material,haindl12ERCIM}, and rendered BTFs from procedurally generated and scanned SVBRDFs. In Figure~\ref{fig:tiling2}, we show examples of the results of NeuBTF for a variety of materials with highly complex structures and reflectance properties, like colored specular (first column) or anisotropy (last). We show some additional results in Figure~\ref{fig:comparisons_qualitative} for materials of different datasets. As shown, our model achieves high quality reconstructions regardless on the type of data source. In Table~\ref{tab:Material}, we show the reconstruction error for the same materials, averaged across the full directional space, for a variety of pixel-wise and perceptual metrics.

\REMOVE{
\Elena{esto de las normales es muy confuso, lo quitaría} \Carlos{Sí? No se, a mi me mola pero me has generado muchas dudas.}We also evaluate whether the encoded materials preserve the geometric properties in their training datasets. We leverage \emph{Photometric Stereo}~\cite{ikeuchi1981determining} to compute surface normals of encoded and ground truth materials. We provide results in Figure~\ref{fig:normals_reconstruction}, where we show that the reconstructed surface geometry is accurate albeit noisier, and preserves the structure of the material. %

\begin{figure}[t!]
	\centering
	\includegraphics[width=1\linewidth]{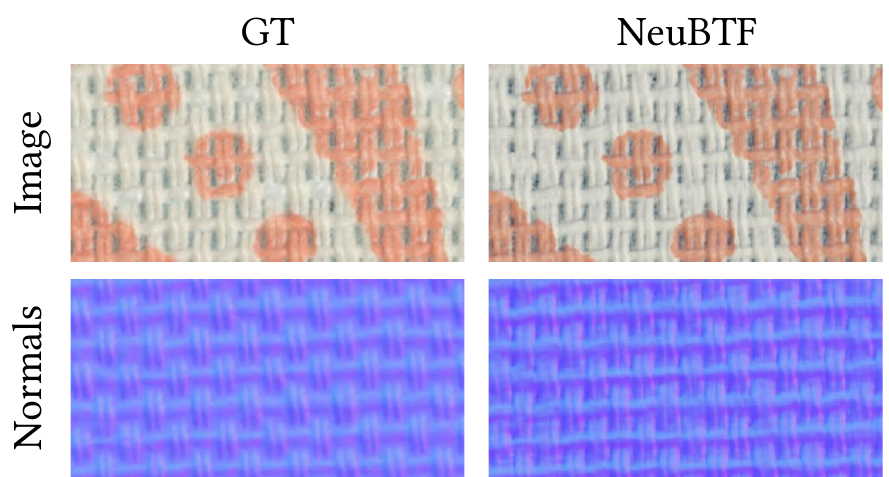}
	\caption{Surface normals estimated using \emph{Photometric Stereo}~\cite{ikeuchi1981determining} for ground truth materials and the materials reconstructed by NeuMat. As shown, the encoded materials preserve geometric information without explicit training. }
	\label{fig:normals_reconstruction}
\end{figure}
}

Finally, in Figure~\ref{fig:latent_channels}, we show a colored visualization for a few channels of the latent neural texture $\mathcal{T}$ found for a variety of materials. Because the values for the neural texture are unbounded, to each channel $c\in \mathcal{T}$, we standardize them to 0 mean and unit variance, and apply a $sigmoid(c) = \frac{1}{e^{1-c} + 1}$ non-linearity to make the maps comparable. Without any explicit training, the models learn to separate distinct parts of the material. For example, the model finds distinct latent spaces for \emph{warp} and \emph{weft} yarns on woven fabrics, or separation between color and geometric patterns. This disentanglement provides clues on why the material propagation is possible, and suggests potential future research directions for fine-grained neural material edition.

\subsection{Compression comparisons with previous work}\label{sec:comparisons}

In Table~\ref{tab:parameters}, we show the number of trainable parameters on the decoders of different neural BTF compression algorithms. As shown, our model is competitive with previous work in terms of trainable parameters. This is achieved as we use more complex loss functions than previous work, which help regularize the models, and because our sinusoidal MLP achieves higher quality reconstructions for natural signals than ReLU MLPs, as shown in~\cite{sitzmann2020implicit}. NeuMIP~\cite{kuznetsov2021neumip} uses smaller MLPs, however, they require an additional decoder for their \emph{neural offset} module, which helps them encode parallax effects (See Figure~\ref{fig:failure_case}), for which our model struggles. Our decoder has one order of magnitude fewer parameters than~\cite{rainer2019neural,rainer2020unified}, however, the method in~\cite{rainer2020unified} provides the benefit of fast encoding of new materials, while ours requires a different model for each new material.

\begin{table}[]
	\centering
	\resizebox{\columnwidth}{!}{%
		\begin{tabular}{@{}r|cccc@{}}
			\toprule
			Method             & \textbf{NeuBTF (ours)} & NeuMIP~\cite{kuznetsov2021neumip}      & Rainer 2019~\cite{rainer2019neural}  & Rainer 2020~\cite{rainer2020unified} \\ \midrule
			Decoder Parameters & \textbf{3011}          & 3332        & 35725       & 38269      \\ \midrule
			Texture Channels   & \textbf{14}            & \textbf{14} & \textbf{14} & 38         \\ \bottomrule
		\end{tabular}%
	}
	\caption{Number of trainable parameters in the decoders and amount of latent texture channels for different neural BTF compression algorithms. We use Torchinfo~\cite{Yep_torchinfo_2020} for this analysis. Note that exact comparisons are challenging, as~\cite{kuznetsov2021neumip} optimizes a multi-level texture pyramid and~\cite{rainer2020unified} learns a latent vector which can encode novel materials. For neither our method nor~\cite{rainer2019neural,rainer2020unified}, we count the parameters in the encoders, as they are not needed for using the materials on rendering systems.}
	\label{tab:parameters}
\end{table}

\REMOVE{
\begin{figure*}[ht!]
	\centering
	\includegraphics[width=1\linewidth]{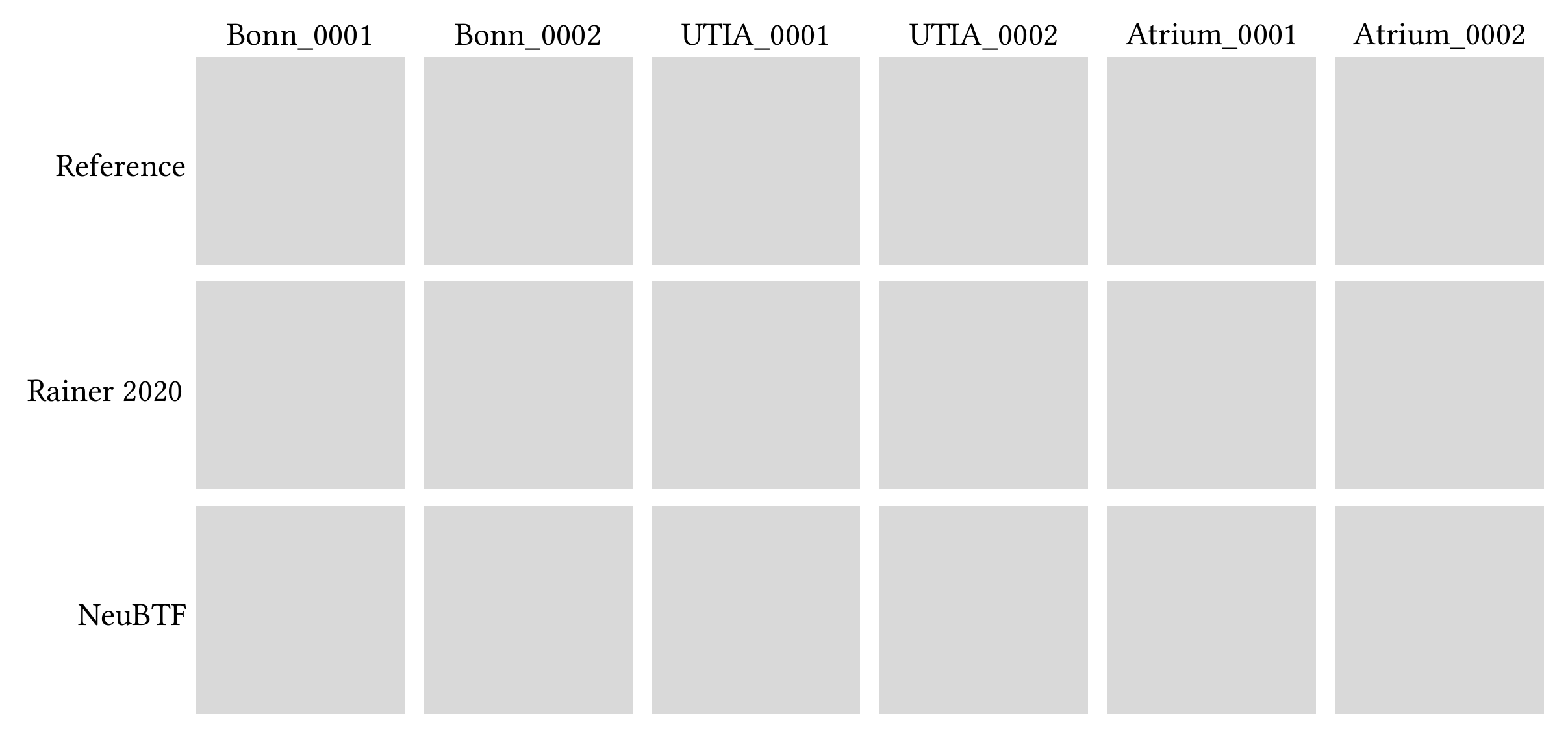}
	\caption{Comparisons of our method with previous work on a variety of real captured BTFs. We observe higher reconstruction quality at a lower parameter count using our method, regardless of the dataset used and across a variety of material types.}
	\label{fig:comparisons}
\end{figure*}}

\REMOVE{
\begin{figure*}[ht!]
	\centering
	\includegraphics[width=1\linewidth]{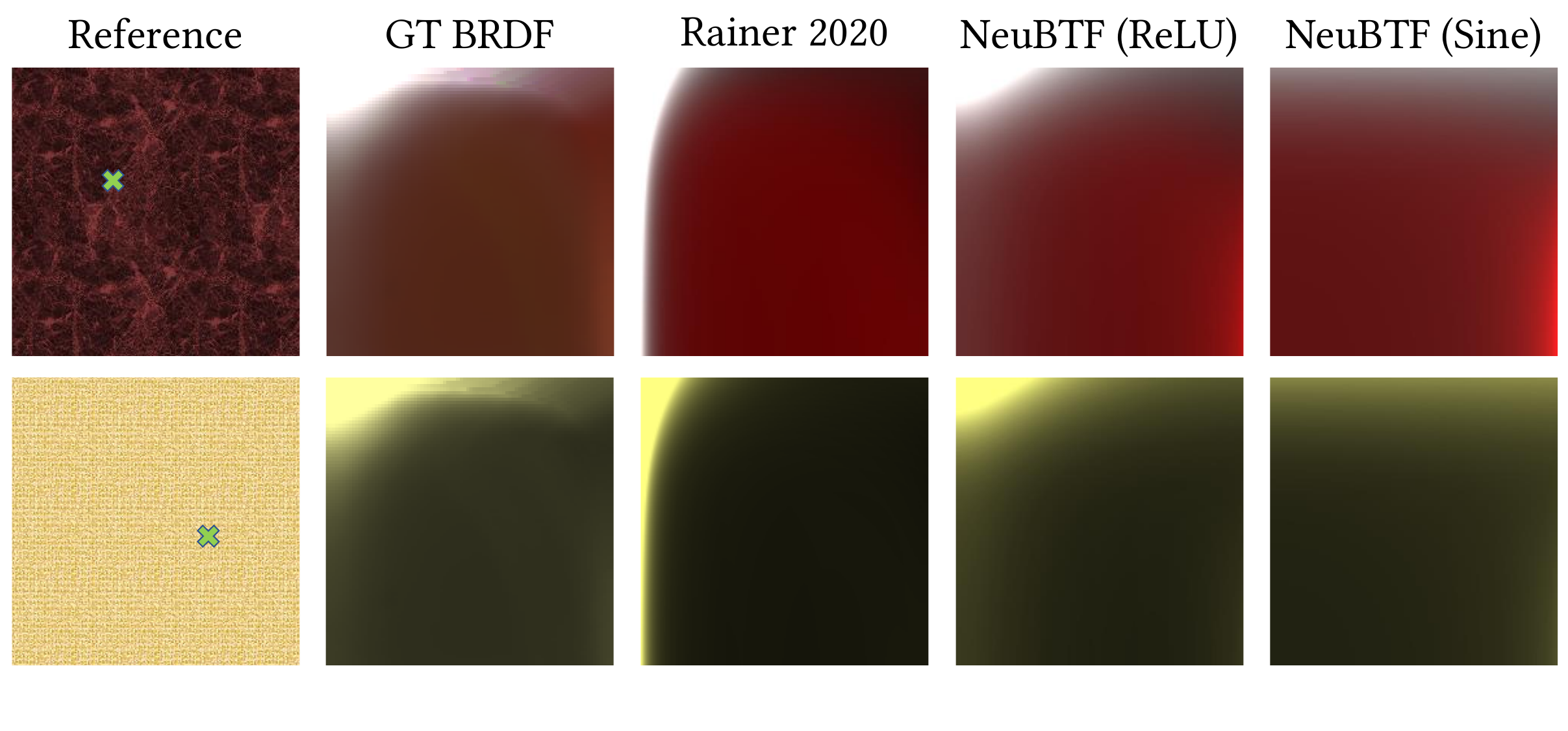}
	\caption{Qualitative comparisons of the local BRDF measurements and the BRDF optimized by Rainier et al \Carlos{cite} and two configurations of our decoder, one with ReLU non-linearities and our final model with leverages sinusoidal activations (rightmost column). We show the location of the queried texel using crosses on the leftmost images. The BRDF slice visualization is provided by the Burley BRDF explorer \Carlos{cite}. \Carlos{WIP, esta todo inventado} }
	\label{fig:comparisons_brdf}
\end{figure*}}

\subsection{Limitations}

\begin{figure}[t!]
	\centering
	\includegraphics[width=1\linewidth]{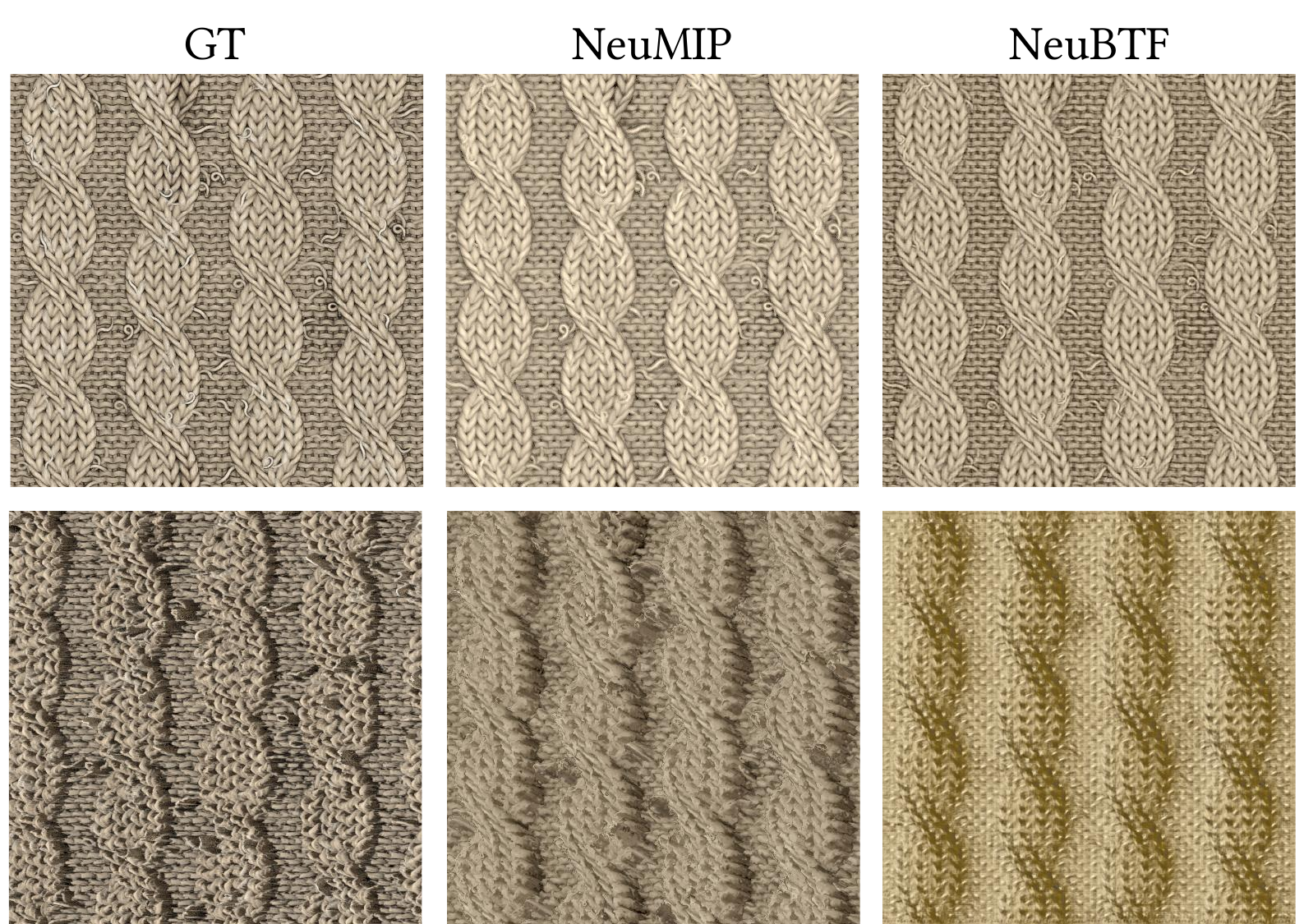}
	\caption{A failure case of our method. Compared to NeuMIP~\cite{kuznetsov2021neumip}, which explictly models parallax effects, our model struggles to accurately encode materials with strong displacements, as this synthetic cable knit from Substance3D. For this type of materials, NeuBTF accurately encodes orthogonal viewing angles (top row), however, it struggles at grazing angles (bottom row).}
	\vspace{-6mm}
	\label{fig:failure_case}

\end{figure}

As we show in Figure~\ref{fig:failure_case}, our model struggles with materials with strong displacement. While our method provides accurate encodings on viewing angles close to the material surface, it cannot accurately encode grazing angles for such extreme cases.  Displacement maps translate the geometric position of the points over the surface, breaking the underlying assumptions behind our neural texture. NeuMIP~\cite{kuznetsov2021neumip} solves this issue by explicitly modelling parallax effects with a \emph{neural offset} module. While we did not observe that such extension was needed for acquired BTF data, like the UBO2014~\cite{schwartz-2014-setups} dataset, introducing a similar module into our editable neural material framework is an interesting future research direction to increase its generality.

\REMOVE{
\begin{itemize}
    \item Limitations shared with BTFs: Materials with holes, transmittance effects, etc.
    \item Comparisons with NeuMIP: Our model struggles with strong parallax effects. Show some examples and propose modifications to the neural representation to mitigate these problems in future work. Also, show 1-2 examples of NeuMIP trained on Bonn BTFs where our model behaves similarly with a comparable number of parameters.
\end{itemize}}

\section{Applications}\label{sec:applications}

\subsection{Reflectance Propagation and Tileable Neural BTFs}
Many material reflectance acquisition devices are limited in the surface dimensions they can digitize. This hinders their applicability to many real-world materials, which exhibit variations that cannot be captured at such small scales. Further, in many applications like SVBRDF acquisition, obtaining larger samples of the material improves realism and helps tileable texture synthesis. In this context, previous work on BTF reflectance compression inherit the surface area limitations of the capture devices used to generate their training data. Our method can easily be applied for reflectance propagation. We build upon the work of Rodriguez-Pardo and Garces~\cite{rodriguezpardo2021transfer} and leverage our \emph{encoder} to propagate the neural texture optimized using a small portion of the material (e.g. a $1\times1$ cm capture) to a larger portion of the same material, represented with a \emph{guidance image} captured using a commodity device like a flatbed scanner. Because our model is trained using a large amount of lighting conditions, as in~\cite{rodriguezpardo2021transfer}, the propagation is invariant to how the images are illuminated. We show results of such pipeline in Figure~\ref{fig:tiling2}. For instance, on the last row, we show an anisotropic and specular \textsc{Silver Jacquard} fabric, for which we generated a BTF by rendering a $1\times1$ cm SVBRDF. This small crop cannot represent the complex pattern in the fabric, which we show on the \emph{guidance image}, which covers a $10\times10$ cm area. Using our encoder, we propagate the neural material to this guidance image, generating a new, high-resolution, latent space which we can render, enabling realistic material representations with a reduced digitization cost. This propagated neural material has a $2000\times2000$ texels resolution, and it requires no re-training during test time.

\REMOVE{
\begin{figure}[t!]
	\centering
	\includegraphics[width=1\linewidth]{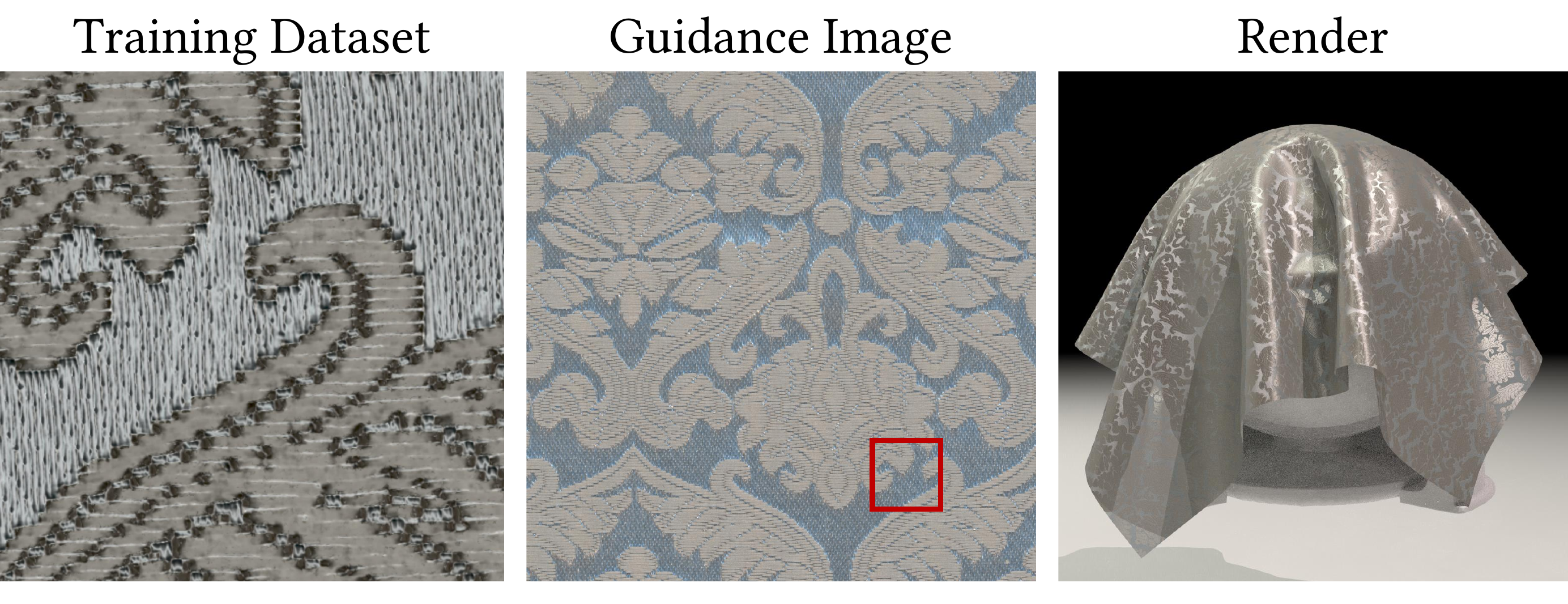}
	\vspace{-5mm}
	\caption{An application of our system for propagating BTFs. On the left, we show the training dataset, which is a $1\times1$ cm portion of a complex \textsc{Silver Jacquard} fabric. We use a flatbed scanner to digitize a \emph{guidance image} which represents a larger ($10\times10$ cm) area of the material. Using it as input to the autoencoder, we generate a larger neural texture. We highlight the original captured surface area as a red inset. On the right, we show a path traced render of the propagated neural material on a complex geometry, showcasing the  anisotropic and specular behaviour of the fabric.}
	\label{fig:propagation}
\end{figure}
}

Relatedly, our propagation framework can also be easily leveraged for generating tileable BTFs. Given any \emph{guidance image} of the material, we can generate a tileable version of it, either using manual editions by artists or automatic algorithms~\cite{moritz2017texture,li2020inverse,rodriguez2019automatic,rodriguezpardo2022SeamlessGAN}. With this tileable input guidance, we can use our autoencoder $\mathcal{A}$ to propagate the neural texture $\mathcal{T}$, effectively generating tileable BTFs, as we show in Figure~\ref{fig:tiling2}. This propagation algorithm can leverage state-of-the-art algorithms for tileable texture synthesis without any modification of our material model or training framework. Tileable BTFs were not achievable with previous approximations and this simple pipeline has the potential of enabling novel applications of this type of material representation in rendering scenarios.

\subsection{Structural Material Edition}

\begin{figure}[t!]
	\centering
	\includegraphics[width=\linewidth]{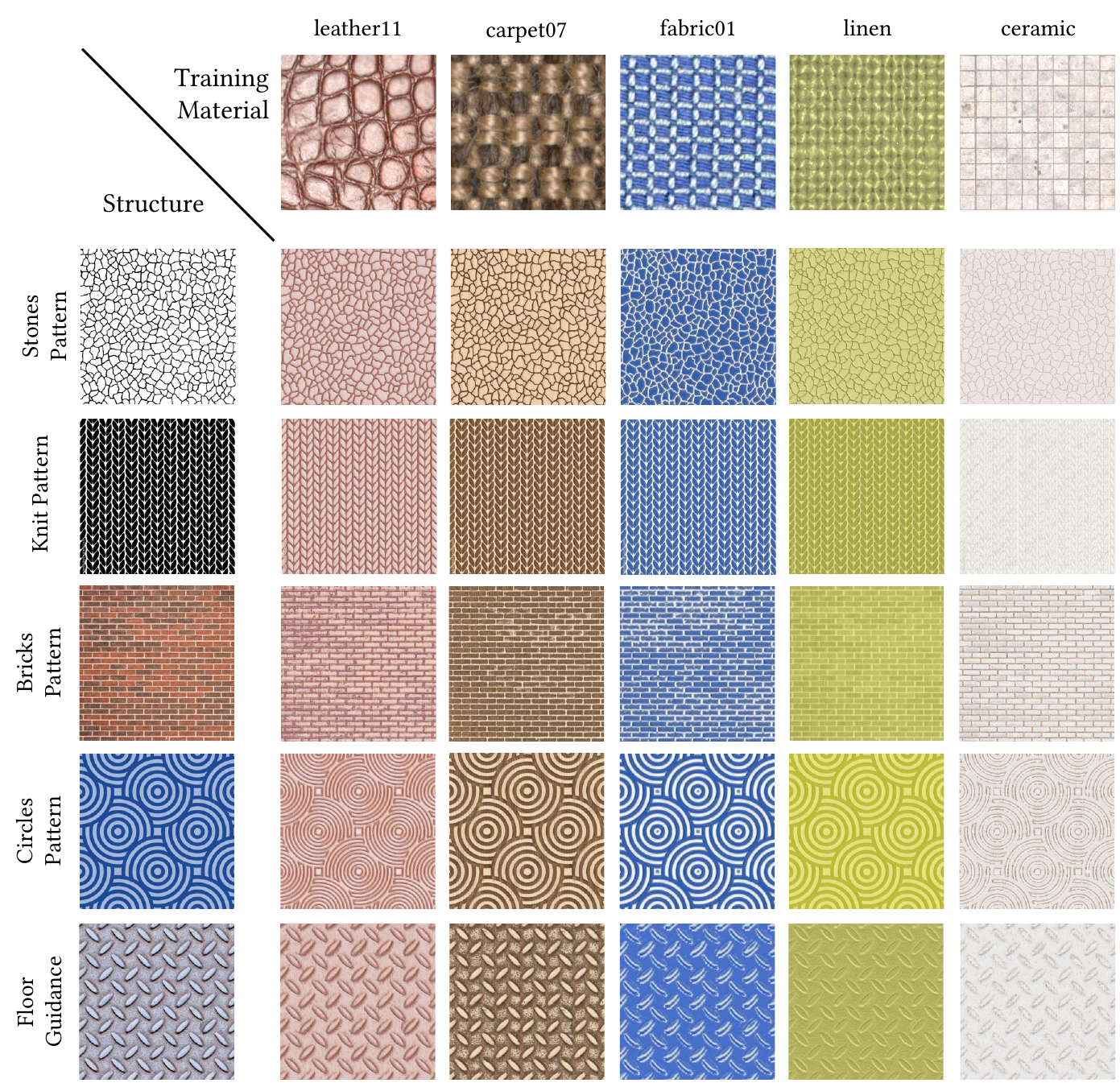}
	\caption{Examples of structural editions allowed by our method on a variety of materials. On the leftmost column, we show \emph{guidance images}, which represent structures into which we transfer the neural BTF measurements illustrated on the top row. As shown, our method can effectively propagate BTF measurements into many material and structure types, using as guidances either synthetic or real images. The first three materials (\emph{leather11, carpet07, fabric01}) are taken from~\emph{UBO 2014}~\cite{schwartz-2014-setups}, the \emph{linen} material is rendered from a captured SVBRDF, while the \emph{ceramic} material is rendered from an artistic material taken from \emph{Substance3D}. We show renders generated using $\theta_v = \theta_l = 0$. Additional results are provided on the supplementary material.}
	\label{fig:structural_editions}
\end{figure}

Besides propagating BTF measurements to larger portions of the material, NeuBTF allows for generating novel materials using structural editions.  Given a trained NeuBTF and a guidance image representing some particular target structure, we can propagate the neural texture to this guidance image, generating high-quality neural materials which preserve the structure of the guidance image and the reflectance properties of the trained neural material. This pipeline allows for easily generating multiple different neural materials without the need for retraining. As we show in Figure~\ref{fig:structural_editions}, this propagation method works for many types of input guidances, including vector black and white images, procedurally generated textures, or real photographs of materials and textures. We show results on acquired BTFs and from synthetic BTFs, rendered from scanned and manually generated SVBRDFs. As shown, our propagation frameworks provides high-quality material editions, even for very challenging cases, like the \emph{circles pattern}.

\subsection{Multi Resolution Neural Materials} \label{sec:lod}
\begin{figure}[t!]
	\centering
	\includegraphics[width=1\linewidth]{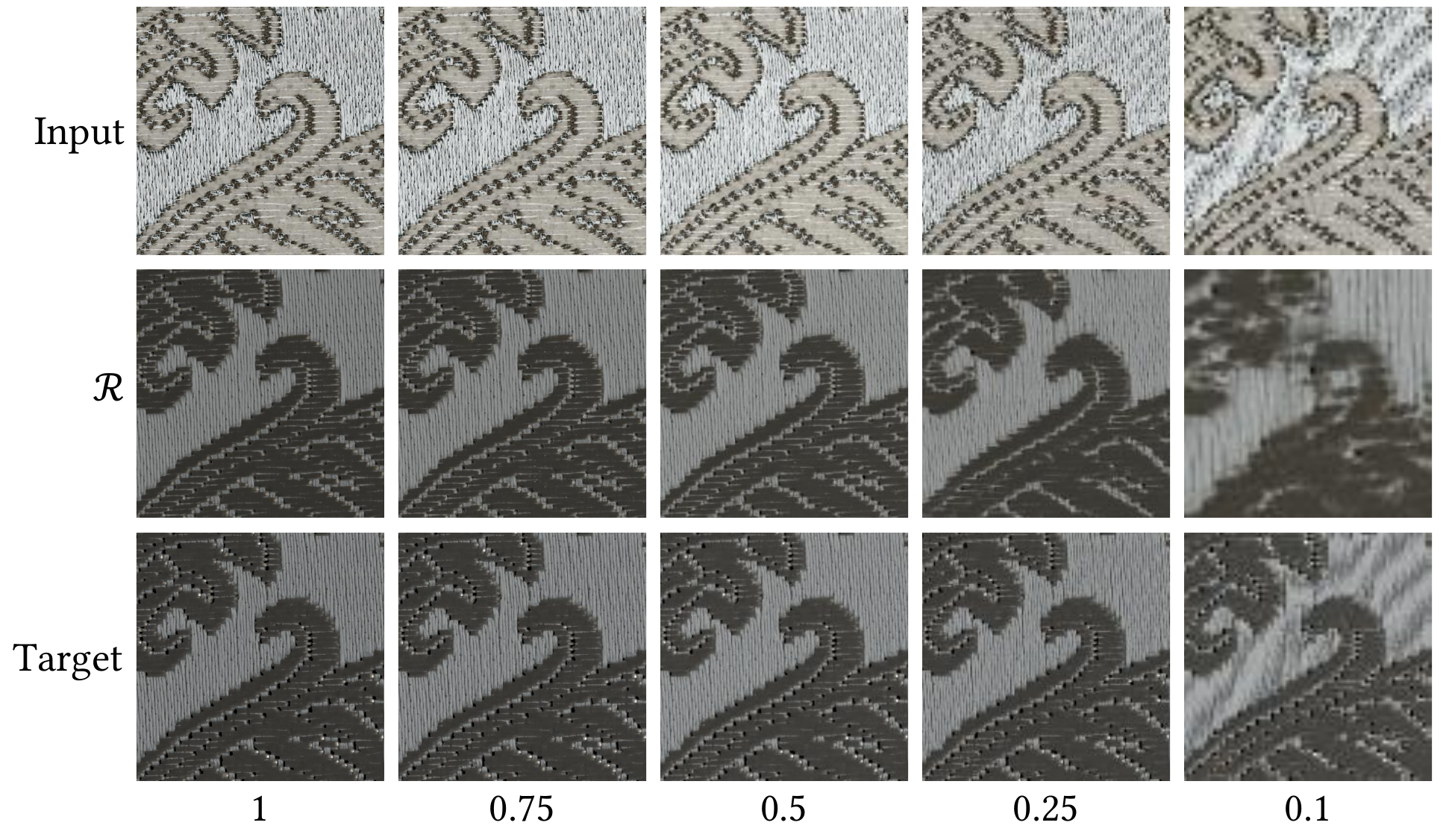}
	\caption{Our method naturally enables for the generation of different resolutions for the neural materials. We show the input to the autoencoder (top row), the rendered material at $\theta_c=75, \theta_l=60, \phi_c =\phi_l=0$ (middle row), and the ground truth image at those positions (bottom), at different resolutions (columns). We achieve this by downsampling the guidance image $\mathcal{G}$ fed into the autoencoder module, which returns an accurately downsampled latent space, thanks to our data augmentation policy applied during training. The rightmost column lies beyond the ranges in which we train the model, however, the results are still somewhat plausible.  }
	\label{fig:LOD}
\end{figure}

Another useful application enabled by our method is the generation of materials at different resolutions. Unlike previous work~\cite{kuznetsov2021neumip}, which explicitly optimizes a pyramid of levels of detail during training, we can generate materials at any resolution at test time without introducing any additional complexities to our material representation. %
Because we train our models using random rescales as a data augmentation policy, they are equivariant to rescales of its input guidance images $G\downarrow$: $\mathcal{R}(\mathcal{A}(\mathcal{G}))\downarrow = \mathcal{R}(\mathcal{A}(\mathcal{G}\downarrow))$. As such, we can generate any continuous resolution for a particular BTF by downsampling the guidance image to the target resolution and propagating its neural texture, as we illustrate in Figure~\ref{fig:LOD}. Note that this algorithm only guarantees accurate results for the rescaling ranges that we use during data augmentation.

\section{Conclusions}\label{sec:conclusions}
We have presented a learning based representation for material reflectance which provides efficient encoding and powerful propagation capabilities. Our method introduces input conditioning into neural BTF representations. This allows for multiple applications which were not possible with previous neural models, including BTF extrapolation, tiling and novel material synthesis through structure propagation. Our method builds upon recent work on neural fields, network design and data augmentation, showing competitive compression capabilities with previous work on neural BTF representation. Through multiple analyses, we have shown the capabilities of our method on a variety of materials with different reflectance properties, including anisotropy or specularity, as well as effectively handling either synthetic and acquired BTFs. 

Our method can be extended in several ways. The most immediate extension is to allow for materials with strong parallax effects due to displacement mapping or curvature, as in~\cite{kuznetsov2021neumip,kuznetsov2022rendering}. Our representation is limited to opaque materials. Extending them to handle translucent or holed surfaces would increase their realism in materials like thin fabrics or meshes. Further, our method could be extended to allow for hyperspectral BTF data~\cite{rump2010groundtruth}, but captured data is scarce.
Besides, recent work on neural BRDF representations~\cite{sztrajman2021neural,rainer2022neural,Fan:2022:NLBRDF} and generative models~\cite{muller2019neural} suggests a promising research direction: Learning to sample from neural BTFs, using invertible neural networks. While these may introduce challenging complexities to the models, they could provide efficient representations for Monte Carlo rendering using importance sampling. 
Further, building upon recent work on SVBRDF capture~\cite{rodriguezpardo2023UMat,zhou2022tilegen}, BRDF sampling~\cite{nielsen2015optimal} and BTF compression~\cite{rainer2020unified}, it could be possible to learn a prior over neural BTFs with a generative model. This should help in capturing more efficiently the data needed for generating these assets, as well as generating new materials and interpolating between them.
Finally, editing semantic and reflectance properties in neural fields is an active area of research~\cite{wu2022palettenerf, liu2021editing, wang2022clip, ye2022intrinsicnerf,instructnerf2023}.  While our method introduces structural edition into neural BTF representations, it is not capable of editing particular semantic properties, such as albedo or specularity. Extending our edition capabilities to more fine-grained parameters is an interesting research avenue. We hope our method inspires future research on neural material representations.

\paragraph*{Acknowledgments}
Elena Garces was partially supported by a Juan de la Cierva - Incorporacion Fellowship (IJC2020-044192-I). This publication is part of the project TaiLOR, CPP2021-008842 funded by MCIN/AEI/10.13039/501100011033 and the NextGenerationEU / PRTR programs.

\bibliographystyle{cag-num-names}
\bibliography{refs}

\end{document}